\documentclass[conference]{IEEEtran}
\IEEEoverridecommandlockouts

\usepackage{times}
\usepackage{helvet}
\usepackage{courier}
\usepackage{color}
\usepackage{amsmath}
\usepackage{algorithm}
\usepackage{algorithmic}
\usepackage{booktabs}
\usepackage{diagbox}
\usepackage{stackengine} 
\newcommand\oast{\stackMath\mathbin{\stackinset{c}{0ex}{c}{0ex}{\ast}{\bigcirc}}}

\usepackage{graphicx}
\newcommand{\tabincell}[2]{\begin{tabular}{@{}#1@{}}#2\end{tabular}}
\usepackage{epsfig,amsfonts,amssymb,times,subfigure,floatflt,wrapfig,enumerate,tikz,url,bm}
\usepackage{siunitx}
\newtheorem{example}{Example}

\newtheorem{definition}{Definition}

\newcommand{\mat}[1]{\bm{#1}}
\newcommand{\ten}[1]{\bm{\mathcal{#1}}}

\floatname{algorithm}{Procedure}

\def\BibTeX{{\rm B\kern-.05em{\sc i\kern-.025em b}\kern-.08em
    T\kern-.1667em\lower.7ex\hbox{E}\kern-.125emX}}
\begin{document}

\title{HOTCAKE: Higher Order Tucker Articulated Kernels for Deeper CNN Compression
}
\author{Rui Lin$^{1}$,
Ching-Yun Ko$^{2}$,
Zhuolun He$^{3}$,
Cong Chen$^{4}$,
Yuan Cheng$^{5}$,
Hao Yu$^{6}$
Graziano Chesi$^{7}$,
Ngai Wong$^{8}$,
\\
\\
$^{1}$rlin@eee.hku.hk, 
$^{2}$cyko@mit.edu,
$^{3}$zleonhe@gmail.com,
$^{4}$chencong@eee.hku.hk, \\
$^{5}$cyuan328@sjtu.edu.cn,
$^{6}$yuh3@sustc.edu.cn
$^{7}$chesi@eee.hku.hk,
$^{8}$nwong@eee.hku.hk,
}

\maketitle

\begin{abstract}
The emerging edge computing has promoted immense interests in compacting a neural network without sacrificing much accuracy. In this regard, low-rank tensor decomposition constitutes a powerful tool to compress convolutional neural networks (CNNs) by decomposing the 4-way kernel tensor into multi-stage smaller ones. Building on top of Tucker-2 decomposition, we propose a generalized Higher Order Tucker Articulated Kernels (HOTCAKE) scheme comprising four steps: input channel decomposition, guided Tucker rank selection, higher order Tucker decomposition and fine-tuning. By subjecting each CONV layer to HOTCAKE, a highly compressed CNN model with graceful accuracy trade-off is obtained. Experiments show HOTCAKE can compress even pre-compressed models and produce state-of-the-art lightweight networks.
\end{abstract}

\begin{IEEEkeywords}
Convolutional neural network, Higher order Tucker decomposition
\end{IEEEkeywords}

\section{Introduction}
\label{sec:intro}
Deep learning and deep neural networks (DNNs) have witnessed breakthroughs in various disciplines, e.g.,~\cite{krizhevsky2012imagenet,silver2016mastering}. However, the progressively advanced tasks, as well as the ever-larger datasets, continuously foster sophisticated yet complicated DNN architectures. Despite the sentiment that the redundancy of parameters contributes to faster convergence~\cite{hinton2012improving}, over-parameterization unarguably impedes the deployment of modern DNNs on edge devices constrained with limited resources. This dilemma intrinsically highlights the demand of compact neural networks. Mainstream DNN compression techniques roughly comprises three categories, namely, pruning, quantization and low-rank decomposition, as depicted below:\\
\textit{Pruning}-- It trims a dense network into a sparser one, either by cropping the small-weight connections between neural nodes (a.k.a. fine-grained)~\cite{han2015learning}, or by removing entire filters and/or even layers (a.k.a. coarse-grained) via a learning approach~\cite{he2017channel,li2016pruning}.\\
\textit{Quantization}-- It limits network weights and activations to be in low bit-widths (e.g., zero or powers of two~\cite{li2016ternary}) such that expensive multiplications are replaced by cheap shift operations. As extreme cases, binary networks such as BNN~\cite{courbariaux2016binarized} and XNOR-Net~\cite{rastegari2016xnor} use only $1$-bit representation that largely reduces computation, power consumption and memory footprint, but at the cost of sometimes drastic accuracy drop.\\
\textit{Low-rank decomposition}-- Initially, low-rank singular value decomposition (SVD) was performed on fully connected (FC) layers~\cite{denton2014exploiting}. It was later recognized that filters of a CONV layer can be aggregated as a $4$-way kernel (filter) tensor $(\text{height} \times  \text{width} \times \text{\#inputs} \times \text{\#outputs})$ and decomposed into low-rank factors for compression. For example, Ref.~\cite{lebedev2014speeding} uses canonical polyadic (CP) decomposition to turn a CONV layer into a sequence of four convolutional layers with smaller kernels. However, this approach only compresses one or several layers instead of the whole network. Its manual rank selection also makes the procedure time-consuming and ad-hoc. Ref.~\cite{Kim2016CompressionOD} overcomes this by utilizing Tucker-2 decomposition to factorize a CONV layer into three successive stages of smaller kernels, whose corresponding Tucker ranks are searched via Variational Bayesian Matrix Factorization (VBMF). Ref.~\cite{hayashi2019einconv} surveys various tensor decompositions and their use in compressing CONV layers empirically. However, all these works invariably adopt a $4$-way view of the convolutional kernel tensor.


This work is along the line of tensor decomposition, and recognizes the unexploited rooms for deeper compression by going beyond $4$-way. Specifically, we show, for the first time, that it is possible to further tensorize the \#inputs axis into smaller modes, and as a result achieve higher compression with a tolerable accuracy drop. Our key contributions are:
\begin{itemize}
\item We lift the $4$-way bar of viewing a CNN kernel tensor and relax the Tucker-2 decomposition~\cite{Kim2016CompressionOD} to arbitrary orders. We subsequently propose \textbf{H}igher \textbf{O}rder \textbf{T}u\textbf{C}ker \textbf{A}rticulated \textbf{KE}rnels (HOTCAKE) for granular CONV layer decomposition into smaller kernels and potentially higher compression.
\item Although VBMF provides a principled way of Tucker rank selection, it does not guarantee a global or locally optimal combination of ranks. To this end, we adapt the rank search in a neighborhood centered around VBMF-initialized ranks. Such finite search space largely alleviates the computation of traditional grid search and locates a locally optimal combination of Tucker ranks that work extremely well in practice.
\item HOTCAKE is tangential to other compression techniques and can be applied together with pruning and/or quantized training etc. Being a generic technique, it can be applied to all CNN layers, which is crucial since nowadays fully convolutional networks (FCNs)~\cite{shelhamer2017fully} are widely used in autonomous driving and robot navigation etc. 
\end{itemize}

Experimental results on some state-of-the-art networks then demonstrate that HOTCAKE produces models that strike an elegant balance between compression and accuracy, even when compressing a pre-compressed neural network. In the following, Section~\ref{sec:tensors} introduces some tensor basics. Section~\ref{sec:hotcake} introduces HOTCAKE. Section~\ref{sec:results} presents the experimental results and Section~\ref{conclude} concludes the paper.
\section{Tensor Basics}
\label{sec:tensors}
\begin{figure}[t] 
\begin{center} 
\includegraphics[width=0.4\textwidth]{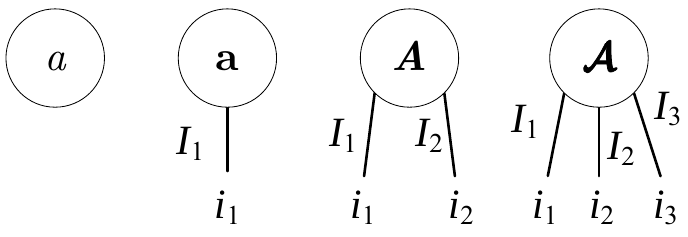}
\caption{Graphical representation of a scalar $a$, vector $\mat{a}$, matrix $\mat{A}$, and third-order tensor $\ten{A}$.}
\label{fig:TNdiagram}
\end{center}
\end{figure}
Tensors are multi-way arrays that generalize vectors (viz. one-way tensors) and matrices (viz. two-way tensors) to their higher order counterparts~\cite{tensorreview}. Henceforth, scalars are denoted by Roman letters $a, b, \ldots$; vectors by boldface letters $\mat{a}, \mat{b}, \ldots$; matrices by boldface capital letters $\mat{A}, \mat{B} \ldots$ and tensors by boldface capital calligraphic letters $\ten{A},\ten{B},\ldots$. Figure~\ref{fig:TNdiagram} shows the so-called \textit{tensor network diagram} for these data structures where an open edge or ``leg'' stands for an index axis. For a $d$-way tensor $\ten{A} \in \mathbb{R}^{I_1 \times I_2 \times I_3 \cdots \times I_d}$, $\ten{A}(i_1, i_2, $ $\ldots, i_d)$ denotes an entry, $i_k$ ($1 \leq i_k \leq I_k$) is the index on the $k$ mode with a dimension $I_k$. A fiber of tensor $\ten{A}$ is obtained by fixing all indices but one. For example, we get a $k$-mode fiber $\ten{A}(i_1,\ldots,i_{k-1},:,i_{k+1},\ldots,i_d) \in \mathbb{R}^{I_k}$ by fixing all other mode indices and scanning through $i_k$. In other words, fibers are high-dimensional analogues of rows and columns in matrices. Employing $numpy$-like notation, ``$\textrm{reshape}(\ten{A},[m_1,m_2,$ $\ldots,m_p] )$'', a $d$-way tensor $\ten{A}\in \mathbb{R}^{I_1 \times I_2 \times I_3 \cdots \times I_d}$ is reshaped into another tensor with dimensions $m_1$, $m_2$, $\ldots, m_p$ that satisfies $\prod_{k=1}^p m_k=\prod_{k=1}^d I_k$. 
Tensor \textit{permutation} rearranges the mode ordering of tensor $\ten{A}$, while keeping the total number of tensor entries unchanged.
Tensor-matrix multiplication or \textit{mode product} is a generalization of matrix-matrix product to that between a matrix along one mode of a tensor:
\begin{definition}(\textbf{$k$-mode product})
The $k$-mode product of tensor $\ten{G} \in \mathbb{R}^{R_1 \times \cdots \times R_d}$ with a matrix $\mat{U} \in \mathbb{R}^{J \times R_k}$ is denoted $\ten{A} = \ten{G} \times_k \mat{U}$ and defined by 
\begin{align*}
\small \ten{A}(r_1, \cdots, r_{k-1}, j, r_{k+1}, \cdots, r_d) &= \hfill \\
\small \sum_{r_k=1}^{R_k} \mat{U}(j, r_k) \ten{G}(r_1, \cdots, r_{k-1}, r_k, &r_{k+1}, \cdots, r_d)
\end{align*}
where $\ten{A} \in \mathbb{R}^{R_1 \cdots R_{k-1} \times J \times R_{k+1} \cdots \times R_d}$.
\end{definition}

With these definitions, the full multilinear product~\cite{cichocki2015tensor} of a $d$-way tensor and $d$ matrices quickly follows:
\begin{definition}(\textbf{Full multilinear product})
The full multilinear product of a tensor $\ten{G} \in \mathbb{R}^{R_1 \times \cdots \times R_d}$ with matrices $\mat{U}^{(1)}, \mat{U}^{(2)}, \ldots, \mat{U}^{(d)}$, where $\mat{U}^{(k)} \in \mathbb{R}^{I_k \times R_k}$, is defined by $\ten{A} = \ten{G} \times_1 \mat{U}^{(1)} \times_2 \mat{U}^{(2)} \ldots \times_d \mat{U}^{(d)}$, where $\ten{A} \in \mathbb{R}^{I_1 \times \ldots \times I_d}$.
\end{definition}

Now the Tucker decomposition follows:
\begin{definition}(\textbf{Tucker decomposition})
Tucker decomposition represents a $d$-way tensor $\ten{A} \in \mathbb{R}^{I_1 \times \ldots \times I_d}$ as the full multilinear product of a core tensor $\ten{G} \in \mathbb{R}^{R_1 \times R_2 \times \ldots \times R_d}$ and a set of factor matrices $\mat{U} \in \mathbb{R}^{I_k \times R_k}$, for $k=1,2,\ldots,d$. Writing out $\mat{U}^{(k)} = [\mat{u}_1^{(k)}, \mat{u}_2^{(k)}, \ldots, \mat{u}_{R_k}^{(k)}]$ for $k = 1,2,\ldots,d$,
\vspace{-0.45em}
\begin{align*}
\label{eqn:tucker1}
\small\ten{A}&=\small\sum\limits^{R_1}_{r_1=1}\cdots\sum\limits^{R_d}_{r_d=1}\ten{G}(r_1,\ldots,r_d)(\mat{u}_{r_1}^{(1)}\circ\cdots\circ\mat{u}_{r_d}^{(d)})\\
\small\nonumber&= \small\ten{G}\, {\times_1}\, \bm{U}^{(1)} \, {\times_2}\, \bm{U}^{(2)} \cdots {\times_d}\, \bm{U}^{(d)}
\end{align*}
where $r_1,r_2,\ldots,r_{d}$ are auxiliary indices that are summed over, and $\circ$ denotes the outer product. 
\end{definition}

The dimensions $(R_1, R_2, \ldots, R_d)$ are called the \textit{Tucker ranks}. We call $\text{rank}(\mat{A}_{(k)})$ the multilinear rank, and $R_k$ is in general no bigger than it. Analogous to SVD truncation, the $R_k$'s can be truncated yielding a Tucker approximation to the original (full) tensor $\ten{A}$.

\section{HOTCAKE}
\label{sec:hotcake}
\begin{figure}[t] 
\begin{center} 
\includegraphics[width=0.48\textwidth]{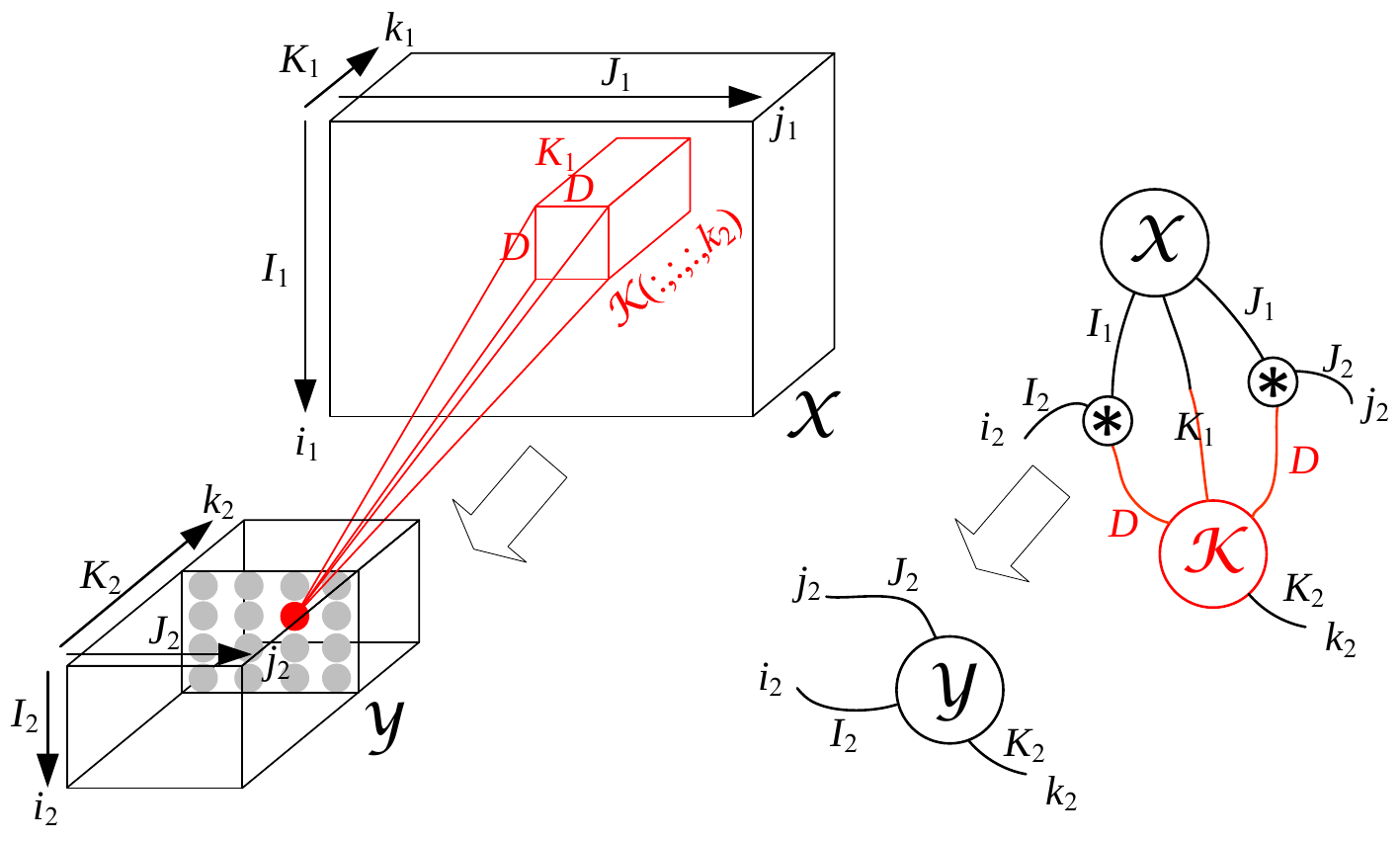}
\caption{Convolution with the input tensors and the kernels.}
\label{fig:conv}
\end{center}
\end{figure}
\begin{figure}[ht] 
\begin{center} 
\includegraphics[width=0.49\textwidth]{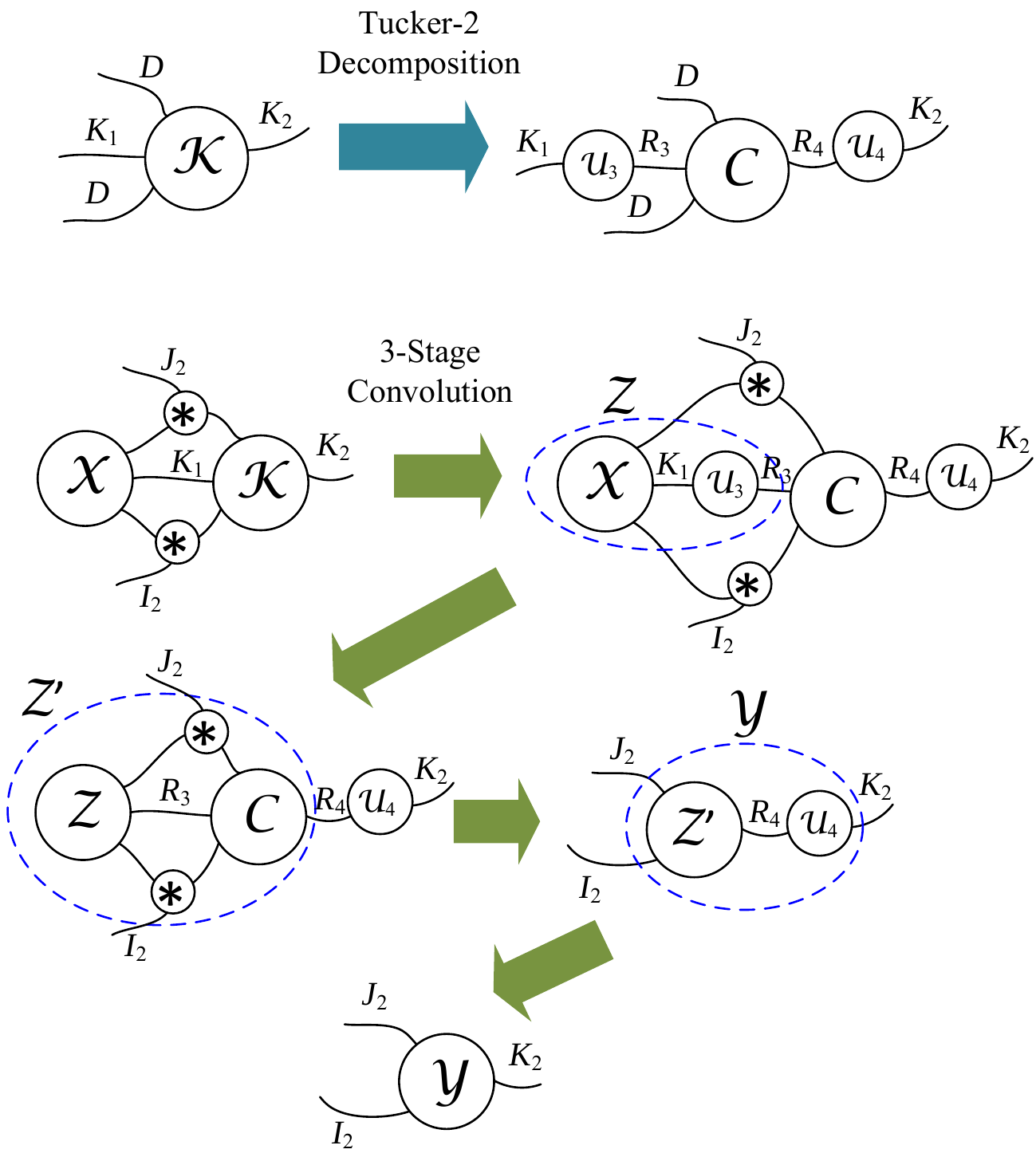}
\caption{(Upper) Tucker-2 decomposition of kernel tensor and (Lower) the three successive, smaller size convolutions marked by blue dashed circles. Some obvious dimensions are omitted in the figure for brevity.}
\label{fig:samsung}
\end{center}
\end{figure}
In each CNN layer, the convolutional kernels form a $4$-way tensor $\ten{K} \in \mathbb{R}^{D \times D \times K_1 \times K_2}$, where $D\times D$ are the spatial dimensions, whereas $K_1$ and $K_2$ are the numbers of input and output channels, respectively. Figure~\ref{fig:conv} illustrates through tensor network diagram how convolution is done via a particular kernel (filter) producing the $k_2$th slice in the output tensor (a.k.a. feature map). Specifically, a CNN filter $\ten{K}(:,:,:,k_2) \in \mathbb{R}^{D \times D \times K_1}$ strides across the input tensor $\ten{X}\in \mathbb{R}^{I_1 \times J_1 \times K_1}$ in the spatial dimensions to produce the $k_2$th slice in the output tensor $\ten{Y} \in \mathbb{R}^{I_2 \times J_2 \times K_2}$. In the tensor network diagram, we adopt the convolution symbol $\oast$ from~\cite{hayashi2019einconv} that abstracts the convolution along the input spatial dimensions to produce the corresponding output spatial dimensions, with stride and zero padding implicitly considered. Such notation hides the unnecessary details and captures the CNN flow clearly, thereby further justifying the benefits of tensor network diagrams.

The basis of HOTCAKE arises from the work in~\cite{Kim2016CompressionOD} that performs Tucker-$2$ decomposition on the mode-$3$ (\text{\#inputs}) and mode-$4$ (\text{\#outputs}) of the kernel tensor $\ten{K}$ so as to decompose a convolutional layer into three smaller consecutive ones. Referring to the upper part of Figure~\ref{fig:samsung}, $(R_3, R_4)$ are Tucker ranks of mode-$3$ and mode-$4$, respectively, and the kernel filters of the three smaller convolution layers are in order $\ten{U}_3 \in \mathbb{R}^{1 \times 1 \times K_1 \times R_3}$, $\ten{C} \in \mathbb{R}^{D \times D \times R_3 \times R_4}$ and $\ten{U}_4 \in \mathbb{R}^{1 \times 1 \times R_4 \times K_2}$. The dashed circles in the lower part of Figure~\ref{fig:samsung} depicts the $3$-stage linear convolution that replaces the original convolution before the nonlinear activation. Contrasting with the right of Figure~\ref{fig:conv} where there are two legs with the convolution operator $\oast$ in which it is assumed $D>1$ (say, $3$ or $5$), such legs are not needed for $D=1$ as in the first and third stages since the input and output feature maps share the same spatial ``legs'' due to the $1\times 1$ nature. Consequently, instead of the one-step $\ten{Y}=\ten{X}\oast \ten{K}$, it becomes $\ten{Z}=\ten{X}\oast \ten{U}_3$,  $\ten{Z}'=\ten{Z}\oast \ten{C}$ and $\ten{Y}=\ten{Z}' \oast \ten{U}_4$. And the number of kernel parameters goes from $D^2K_1K_2$ to $K_1R_3+D^2R_3R_4+K_2R_4$ with the latter being much smaller when $R_3$ and $R_4$ are small.

In short, HOTCAKE innovates by adopting an arbitrary-order perspective of the kernel tensor and allows higher order Tucker decomposition that results in even more granular linear convolutions through a series of small-size articulated kernels (and thereby its name). Moreover, other engineering improvements are incorporated to streamline HOTCAKE into a four-stage pipeline: (1) input channel decomposition; (2) Tucker rank selection; (3) higher order Tucker decomposition; (4) fine-tuning. Their details are described below.

\subsection{Input channel decomposition}

\begin{example}
Suppose a convolution layer of kernel tensor $\ten{K} \in \mathbb{R}^{3 \times 3 \times 128 \times 256}$. In this case, the number of input channels is $K_1 = 128$, which can be decomposed into several branches of dimensions $K_{1i}$'s such that $K_1 = \prod_i K_{1i}$, such as $K_{12} = 16$ and $K_{11} = 8$. These $K_{1i}$'s can be determined according to the estimated number of clusters of filters. Empirically, it is found that it works best when $K_{1j} \geq K_{1i}$, $\forall j \geq i$.
\end{example}
After decomposing the input channels into $l$ branches, the $4$-way kernel tensor $\ten{K} \in \mathbb{R}^{D \times D \times K_1 \times K_2}$ is reshaped into a $(3+l)$-way tensor $\ten{K}_{new} \in \mathbb{R}^{D \times D \times K_{11} \times \ldots \times K_{1l} \times K_2}$. The reason that we do not factorize the \#outputs axis is that the CONV layers are always followed by batch normalization and/or pooling. Such operations are generally done on a $3$-way output tensor, and if we tensorize the \#outputs axis into multi-way, we will need to reshape them back into one mode which increases the computation complexity. Also, Tucker decomposition is not carried out on the spatial dimensions $D\times D$ as they are inherently small, e.g., $D=1,3,5$.

\subsection{Tucker rank selection}
Following from above, the rank-$(R_{3i}, \ldots, R_{3l}, R_4)$ determines the trade-off between the compression and accuracy loss. A manual search of the ranks is time-consuming and does not guarantee appropriate ranks, whereas exhaustive grid-search guarantees the best combo but is prohibitive due to exponential combinations. To estimate the proper ranks, Ref.~\cite{Kim2016CompressionOD} employs the analytic Variational Bayesian Matrix Factorization (VBMF) that is able to find the variance noise and ranks, and thus offers a good yet sub-optimal set of Tucker ranks. To this end, HOTCAKE uses the VBMF-initialized ranks to center the search space and evaluates the rank combos in a finite neighborhood.

\begin{example} Given a kernel tensor $\ten{K} \in \mathbb{R}^{3 \times 3 \times 128 \times 256}$, suppose the input channel decomposition makes it a $\ten{K}_{new} \in \mathbb{R}^{3 \times 3 \times 8 \times 16 \times 256}$ by decomposing its \#inputs axis into $2$ branches. Assuming selected VBMF ranks of $\ten{K}_{new}$ being $(R_{31}, R_{32}, R_4) = (5, 7, 107)$ and a search diameter of $3$, the rank search space in our algorithm is then $\{(R_{31}, R_{32}, R_4) | [4,5,6] \times [6,7,8] \times [106, 107 ,108] \}$, containing $27$ different combinations. 
\end{example} 

Compared with grid-search, our rank selection strategy searches a much smaller finite space and requires much lower computation. It also outweighs a pure VBMF solution, as searching within a region rather than sticking to a point gives a higher possibility to locate a better rank setting. 
Notably, we introduce randomized SVD (rSVD)~\cite{halko2011finding} to the VBMF initialization in replace of conventional SVD to avoid the $\mathcal{O}(n^3)$ computational complexity, where $n$ is the $\max (\# rows, \# columns)$\footnote{Since the input channels are decomposed into several branches, the flattened matrix of tensor $\ten{K}_{new}$ needed for VBMF usually has a very large $\# columns$, leading to the failure of SVD. In contrast, rSVD overcomes this problem by randomly projecting the original large matrix onto a much smaller subspace, while producing practically same results in all our experiments.}.

\subsection{Higher order Tucker compression}
The truncated higher-order singular value decomposition (HOSVD)~\cite{de2000multilinear} and the higher-order orthogonal iteration algorithm (HOOI)~\cite{de2000best} are two widely used algorithms for Tucker decomposition. Here, we employ HOSVD with rSVD in place of SVD as described before. Procedure~\ref{alg:tHOCompress} describes the modified HOSVD.
\begin{algorithm}[th]
\caption{Modified truncated higher-order singular value decomposition (HOSVD)}
\label{alg:tHOCompress}
\begin{algorithmic}
\REQUIRE Tensor $\ten{K}_{new} \in \mathbb{R}^{I_1 \times \ldots \times I_d}$ , ranks: $R_1, \ldots, R_d$.
\ENSURE Core tensor $\ten{G} \in \mathbb{R}^{R_1 \times \ldots \times R_d}$, factor matrices $\mat{U}^{(1)}, \ldots, \mat{U}^{(d)}$, where $\mat{U}^{(k)} \in \mathbb{R}^{I_k \times R_k}$ for $k=1,\ldots,d$.
\FOR{$n = 1,2,\ldots,d$}
\STATE $[\mat{\mat{L}, \mat{\Sigma}, \mat{R}^T}] \leftarrow$ rSVD decomposition of $\mat{K}_{new(n)}$
\STATE $\mat{U}^{(n)} \leftarrow R_n$ leading left columns of $\mat{L}$
\ENDFOR
\STATE $\ten{G} \leftarrow [[\ten{K}_{new}; \mat{U}^{(1)T}, \ldots, \mat{U}^{(d)T}]]$
\end{algorithmic}
\end{algorithm}
We remark that the computation of each factor matrix $\mat{U}^{(n)}$ $(n=1,\ldots,d)$ is independent, since the input matrix $\mat{K}_{new(n)}$ for rSVD is from the tensor $\ten{K}_{new}$ independently. Thus, Tucker decomposition can be done on selected modes in parallel. For a given tensor $\ten{K}_{new}$, after Tucker decomposition, there are $l$ $1 \times 1$ CONV layers and exactly one CONV layer with the same spatial filter size, stride and zero-padding size as the original CONV layer, but the dimensions of input channels and output channels are smaller. The following example shows how input data are convolved with those CONV layers.
\begin{example} Given a kernel tensor $\ten{K} \in \mathbb{R}^{3 \times 3 \times 128 \times 256}$, suppose its input channels are decomposed into $2$ modes. After Tucker decomposition, there are $4$ tensors, denoted \mbox{$\ten{U}_3 \in \mathbb{R}^{1 \times 1 \times 8 \times 5}$}, \mbox{$\ten{U}_4 \in \mathbb{R}^{1 \times 1 \times 16 \times 7}$}, \mbox{$\ten{C} \in \mathbb{R}^{3 \times 3 \times 35 \times 117}$} and \mbox{$\ten{U}_5 \in \mathbb{R}^{1 \times 1 \times 117 \times 256}$}. Note that the factor matrices $\ten{U}_3, \ten{U}_4, \ten{U}_5$ are tensorized with ``singleton'' axes and regarded as $4$-way convolutional kernels as well. The rest of the flow then follows similarly to Figure~\ref{fig:samsung}, but now with two $1\times 1$ CONV layers due to $\ten{U}_3$ and $\ten{U}_4$, followed by the $3\times 3$ CONV of $\ten{C}$ and ended with another $1\times 1$ CONV of $\ten{U}_5$ to produce $\ten{Y}$. 
\end{example}

Next, we analyze the space and time complexities of a CONV layer in HOTCAKE. For storage, the parameter number is in $\mathcal{O}(D^2R_{3i}^l+lR_{3i}K_{1i}+K_4R_4)$, where $R_{3i}$ and  $K_{1i}$ are the largest values in $R_{31},R_{32},\ldots, R_{3l}$ and $K_{11},K_{12},\ldots, K_{1l}$, respectively. We remark that $K_{1i}$ is exponentially smaller than $K_1$, and $R_{3i}$ is further smaller than $K_{1i}$. Therefore, the overall parameter number in a decomposed layer is much smaller than the original $\mathcal{O}(D^2K_1K_2)$. For time complexity, assuming the output feature height and width are the same as those of the input feature after passing the CONV layer, the time complexity is $\mathcal{O}(M^2(D^2R_{3i}^l+lR_{3i}K_{1i}+K_4R_4))$, where $M$ is the output feature height or width value. Recognizing that the time complexity of the original CONV layer is $\mathcal{O}(M^2D^2K_1K_2)$, a huge computational complexity reduction can be achieved.

\subsection{Fine-tuning}
After the above three stages, the accuracy of the Tucker decomposed model often drops significantly. However, the accuracy can be recovered to an acceptable level (in less than $20$ epochs in all our trials) via retraining.


\section{Experimental Results}
\label{sec:results}
We implement the proposed HOTCAKE processing on three popular architectures, namely, SimpNet~\cite{hasanpour2018towards}, MTCNN~\cite{7553523} and AlexNet~\cite{krizhevsky2012imagenet}. The first two are lightweight networks, while the last one is deeper and contains more redundant parameters. We use \textit{CIFAR-10}~\cite{krizhevsky2009learning} dataset as a benchmark for SimpNet and AlexNet. The datasets used for MTCNN are \textit{WIDER FACE} and \textit{CNN for Facial Point}. All neural networks are implemented with PyTorch, and experiments are run on an NVIDIA GeForce GTX1080 Ti Graphics Card with 11GB frame buffer. We compared HOTCAKE with Tucker-2 decomposition~\cite{Kim2016CompressionOD} but not with CP decomposition~\cite{lebedev2014speeding} because the latter can only be used to compress $1$ or $2$ CONV layers and not applicable to the whole network.

\subsection{Experiments on SimpNet}
We first tested with several lightweight CNNs, which aims to show that HOTCAKE can further remove the redundancy in some intentionally designed compact models. The first lightweight CNN we compressed is SimpNet~\cite{hasanpour2018towards}. This net is carefully crafted in a principled manner and has only 13 layers and $5.48$M parameters, while still outperforming the conventional VGG and ResNet etc. in terms of classification accuracy. Due to its efficient and compact architecture, SimpNet can potentially achieve superior performance in many real-life scenarios, especially in resource-constrained mobile devices and embedded systems.

\begin{figure}[t]
\centering
\includegraphics[scale = 0.42]{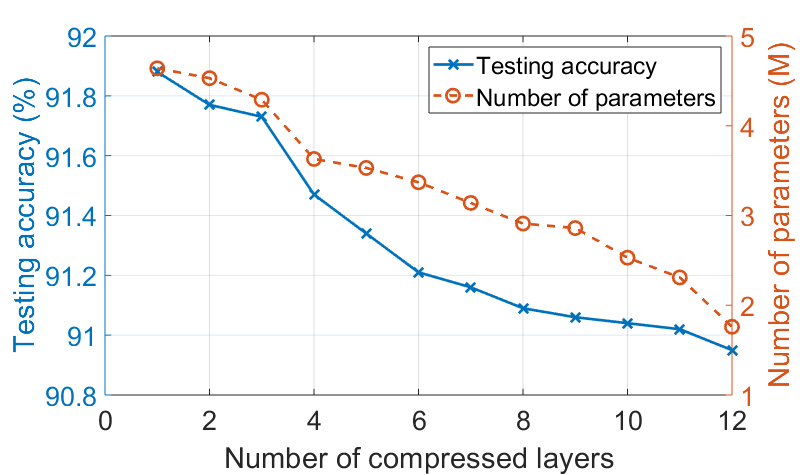}
\caption{Classification accuracy and model parameters vs. the number of compressed CONV layers. We compressed the layers in an order w.r.t. the compression ratios. For example, if only 1 layer is compressed, we choose the $5$th CONV layer since it has the highest compression ratio. If 2 layers are compressed, the $5$th and $12$th layers are chosen, and so on. The solid curve shows how accuracy decreases when more layers are compressed. The dotted curve shows how number of parameters changes.}
\label{fig:my_label}
\end{figure}

There are totally 13 CONV layers in the SimpNet and we do not compress the first layer since the input channel number is only $3$. Table~\ref{tab:simp1} shows the overall result when compressing the $2-13$ CONV layers with Tucker-2 and HOTCAKE. We notice that the two methods achieve similar classification accuracy after fine-tuning, while HOTCAKE produces a more compact model. The detailed parameter number and compression ratio of each CONV layer are enumerated in Table~\ref{tab:simp2}. We observe that HOTCAKE achieves a higher compression ratio than Tucker-2 almost in every CONV layer. Table~\ref{tab:simp2} also provides hints as to which layer is the most compressible and one can better achieve a balance between the model size and the the classification performance. For example, we can decide which layers should be compressed if a specific model size is given. Fig.~\ref{fig:my_label} shows the classification accuracy of the compressed model obtained by employing HOTCAKE when increasing the number of compressed layers. The sequence we compress the CONV layer is determined by their compression ratios listed in Table~\ref{tab:simp2}. The layers with higher compression ratio will be compressed at the beginning. Employing this strategy, we can achieve the highest classification accuracy when the overall model compression ratio is given.

With the successful application on SimpNet, we argue that our proposed compression scheme can handle the already-tiny model better than Tucker-2. In the next experiment, we consider an even smaller CNN model and compress it using HOTCAKE.
\begin{table}[t]
\centering
\caption{An overview of SimpNet's performance and the number of parameters before and after compression.}
\begin{tabular}{cccc}
\hline
& Original & Tucker-2 & HOTCAKE\\
\hline
Testing Accuracy & $95.21\%$ & $90.84\%$ & $90.95\%$\\
Overall Parameters & $5.48$M & $2.24$M & $1.75$M \\
Compression Ratio & —— & $2.45 \times$ & $3.13\times$\\
\hline
\end{tabular}
\label{tab:simp1}
\end{table}
\begin{table}[t]
\centering
\caption{SimpNet's layer-wise analysis. Numbers in brackets are compression ratios compared with the original CONV layers.}
\begin{tabular}{cccc}
\hline
\tabincell{c}{No. of compressed\\ CONV layers}& Original & Tucker-2 & HOTCAKE \\
\hline
 2 & $76$K & $30$K ($2.53 \times$) & $24$K ($3.17 \times$)\\
 3 & $147$K & $61$K ($2.41 \times$) & $39$K ($3.77 \times$)\\
 4 & $147$K & $61$K ($2.41 \times$) & $43$K ($3.42 \times$)\\
 5 & $221$K & $88$K ($2.72 \times$)& $65$K ($3.40 \times$)\\
 6 & $332$K & $136$K ($2.44 \times$) & $103$K ($3.22 \times$)\\
 7 & $332$K & $137$K ($2.42 \times$) & $92$K ($3.61 \times$)\\
 8 & $332$K & $137$K ($2.42 \times$) & $104$K ($3.19 \times$)\\
 9 & $332$K & $135$K ($2.46 \times$) & $112$K ($2.96 \times$)\\
 10 & $498$K & $206$K ($2.42 \times$) & $162$K ($3.07 \times$)\\
 11 & $746$K & $314$K ($2.37 \times$) & $183$K 
 ($4.08 \times$)\\
 12 & $920$K & $371$K ($2.48 \times$) & $257$K ($3.58 \times$)\\
 13 & $1.12$M & $569$K ($1.97 \times$) & $569$K ($1.97\times$)\\
\hline
\end{tabular}
\label{tab:simp2}
\end{table}

\subsection{Experiments on MTCNN}
The second model we tested is MTCNN~\cite{7553523}, which is designed for human face detection. Aiming for real-time performance, each built-in CNN in MTCNN is designed to be lightweight deliberately. Specifically, MTCNN contains three cascaded neural networks called Proposal Network (P-Net), Refinement Network (R-Net) and Output Network (O-Net).
The size of P-Net and R-Net are too small such that we do not have much space to compress them. Therefore, we compress only the O-Net which contains $4$ CONV layers with the kernel tensor of $\ten{K}^{(1)} \in \mathbb{R}^{3 \times 3 \times 3 \times 32}$, $\ten{K}^{(2)} \in \mathbb{R}^{3 \times 3 \times 32 \times 64}$, $\ten{K}^{(3)} \in \mathbb{R}^{3 \times 3 \times 64 \times 64}$ and $\ten{K}^{(4)} \in \mathbb{R}^{3 \times 3 \times 64 \times 128}$. 


We also do not compress the first CONV layer of O-Net due to the same reason as in SimpNet. Table~\ref{tab:mecnn1} shows the overall model compression result employing HOTCAKE. We achieved at least $3\times$ compression ratio on all the three CONV layers even though the original layer sizes are already small enough. Table~\ref{tab:mtcnn2} further illustrates the detailed performance of the compressed model. The face classification accuracy decreases less than $1\%$ compared with the original model, at the same time the loss increment of the three tasks are all negligible.

\begin{table}[t]
\centering
\caption{O-Net's Layer-wise analysis. Numbers in brackets are compression ratios.}
\begin{tabular}{ccc}
\hline
\tabincell{c}{No. of compressed\\ CONV layers} & Original & HOTCAKE  \\
\hline
2 & $18$K & $4$K ($4.50 \times$) \\
3 & $37$K & $8$K ($4.63 \times$) \\
4 & $33$K & $11$K ($3.00 \times$) \\
\hline
\end{tabular}
\label{tab:mecnn1}
\end{table}
\begin{table}[t]
\centering
\caption{Performances of MTCNN before and after compression.}
\begin{tabular}{ccc}
\hline
& Original & HOTCAKE\\ 
\hline
\tabincell{c}{Face Classification\\ Accuracy} & $95.36\%$ & $94.42\%$ \\
\hline
\tabincell{c}{Loss of Face\\ Detection} & $0.648$ & $0.686$  \\
\hline
\tabincell{c}{Loss of\\ Bounding Box} & $0.0137$ & $0.0175$ \\
\hline
\tabincell{c}{Loss of Face\\ Landmarks} & $0.0107$ & $0.0128$ \\
\hline
Total loss & $0.546$ & $0.569$ \\
\hline
\end{tabular}
\label{tab:mtcnn2}
\end{table}

\subsection{Experiments on AlexNet}
\begin{figure}[H] 
\begin{center}
\includegraphics[width=0.42\textwidth]{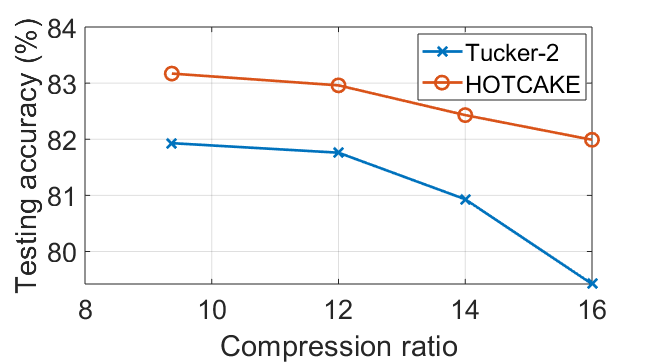}
\end{center}
\caption{Accuracy vs. compression ratio on CIFAR-10.}
\label{fig:HigherRatio}
\end{figure}

The third model we use is AlexNet~\cite{krizhevsky2012imagenet} which is much larger than the above two examples. It contains $61.1$M parameters in total. Again, we compressed all its CONV layers except the first. Table~\ref{tab:alex1} shows the layer-wise analysis of AlexNet. We observe that HOTCAKE can achieve higher compression ratio for each layer. Table~\ref{tab:alex2} further shows classification performance of the compressed models. Tucker-2 obtains a higher accuracy when its compression ratio is half less than HOTCAKE. To make the comparison fair, we further set ranks manually for Tucker-2 to reach the same compression ratio as HOTCAKE, and its classification accuracy drops from $90.29\%$ to $81.39\%$, which is lower than that of HOTCAKE ($83.17\%$). Next, we assign ranks for both Tucker-2 and HOTCAKE, to reach higher compression ratios at around $12 \times$, $14 \times$, and $16 \times$. The results are illustrated in Fig.~\ref{fig:HigherRatio} wherein it is seen that HOTCAKE achieves a higher classification accuracy than Tucker-2 on all the three compression ratios, which indicates the superiority of HOTCAKE over Tucker-2 in high compression ratios. 
\begin{table}[t]
\centering
\caption{AlexNet's layer-wise analysis. Numbers in brackets are compression ratios compared with the original CONV layers.}
\begin{tabular}{cccc}
\hline
\tabincell{c}{No. of compressed\\ CONV layer} & Original & Tucker-2 & HOTCAKE \\
\hline
2 & $307$K & $127$K ($2.42 \times$) & $56$K ($5.48 \times$)\\
3 & $664$K & $197$K ($3.37 \times$) & $120$K ($5.53 \times$)\\
4 & $885$K & $124$K ($7.14 \times$)& $51$K ($17.35 \times$)\\
5 & $590$K & $71$K ($8.31 \times$)& $34$K ($17.35 \times$)\\
\hline
\end{tabular}
\label{tab:alex1}
\end{table}

\begin{table}[H]
\centering
\caption{An overview of AlexNet's performance and number of parameters before and after compression.}
\begin{tabular}{cccc}
\hline
& Original & Tucker-2 & HOTCAKE  \\
\hline
Testing Accuracy & $90.86\%$ & $90.29\%$ & $83.17\%$ \\
\tabincell{c}{Overall Parameters\\ (CONV layers)} & $2.47$M & $520$K & $261$K \\
Compression Ratio & —— & $4.75 \times$ & $9.37 \times$ \\
\hline
\end{tabular}
\label{tab:alex2}
\end{table}

\section{Conclusion}
\label{conclude}
This paper has proposed a general procedure named HOTCAKE for compressing convolutional layers in neural networks. 
We demonstrate through experiments that HOTCAKE can not only compress bulky CNNs trained through conventional training procedures, but it is also able to exploit redundancies in various compact and portable network models. 
Compared with Tucker-2 decomposition, HOTCAKE reaches higher compression ratios with a graceful decrease of accuracy. 
Furthermore, HOTCAKE can be selectively used for specific layers to achieve targeted and deeper compression, and provide a systematic way to explore better trade-offs between accuracy and the number of parameters. 
Importantly, this proposed approach is powerful yet flexible to be jointly employed together with pruning and quantization.
\bibliographystyle{IEEEtran}
\bibliography{dac2020}
\end{document}